\documentclass[letterpaper]{article} 
\usepackage[submission]{aaai24}  
\usepackage{times}  
\usepackage{helvet}  
\usepackage{courier}  
\usepackage[hyphens]{url}  
\usepackage{graphicx} 
\usepackage{natbib}  
\usepackage{caption} 
\usepackage{algorithm}
\usepackage{algorithmic}
\usepackage{multirow}

\usepackage{newfloat}
\usepackage{listings}
\DeclareCaptionStyle{ruled}{labelfont=normalfont,labelsep=colon,strut=off} 
\floatstyle{ruled}
\newfloat{listing}{tb}{lst}{}
\floatname{listing}{Listing}
\title{Seal: Advancing Speech Language Models to be Few-Shot Learners}
\author {
    Shuyu Lei,
    Lingen Liu,
    Jiaolong Yang,
    Yasen Jiao,
    Yuxiang Yang,
    Yushu Yang,
    Xiang Guo
}
\affiliations {
    Didi Chuxing, Beijing, China\\
    leishuyu@didiglobal.com
}

\usepackage{bibentry}

\begin{document}

\maketitle

\begin{abstract}
Existing auto-regressive language models have demonstrated a remarkable capability to perform a new task with just a few examples in prompt, without requiring any additional training. In order to extend this capability to a multi-modal setting (i.e. speech and language), this paper introduces the \textbf{Seal} model, an abbreviation for \textbf{S}p\textbf{e}ech l\textbf{a}nguage mode\textbf{l}. It incorporates a novel alignment method, in which Kullback-Leibler divergence loss is performed to train a projector that bridges a frozen speech encoder with a frozen language model decoder. 
The resulting \textbf{Seal} model exhibits robust performance as a few-shot learner on two speech understanding tasks.
Additionally, consistency experiments are conducted to validate its robustness on different pre-trained language models.
\end{abstract}

\section{Introduction}
\label{intro}
Auto-regressive language models \cite{gpt3} have a very impressive ability to perform a new task with a few examples without any additional fine-tuning, named few-shot learners.
Despite these remarkable capabilities, such language models are deaf to speech modalities. 
Therefore, a separate automatic speech recognition (ASR) model is necessary to transcribe speech into text, enabling language models to act as few-shot learners for speech modal, which suffers from the transcription errors caused by ASR.

This paper aims to enhance the few-shot capabilities of language models for the speech modality.
The foremost challenge of advancing language models to be speech few-shot learners is that aligning speech features into word embedding space of pre-trained language models effectively.
Prior speech language models \cite{salmonn,qwen-audio} utilize next token prediction task to align speech features into word embedding space, where the desired next tokens are generated by task-related prompt and speech features.
These configurations are expressed as $P(o|s,p)$, where $s=[s_1,...,s_l]$ represents speech features, $p=[p_1,...,p_m]$ signifies the task-related prompt, and $o=[o_1,...,o_n]$ denotes the intended output text.
This formulation has a negative impact on context learning, as it overlooks the fact that speech features and their corresponding transcripts should exhibit the same behaviors in pre-trained language models, which should be formulated as $P(o|s,p)=P(o|t,p)$, where the $t=[t_1,...,t_I]$ denotes the transcripts of speech features.

In the research area of mechanistic interpretability, \cite{induction} introduces that the ability of context-learning is derived from an induction head, a circuit whose complete the pattern by copying and completing sequences that have occurred before.
Inspired by induction heads, this paper argue that extending pre-trained language models to function as few-shot learners for speech should be formulated as probabilistic models $P(o|s) = P(o|t)$.
Therefore, this paper introduces a speech language model that consists of three parts: a frozen speech encoder, a trainable projector, and a frozen language model, where the trainable projector maps the speech features to the word embedding space of language models.
The projector is trained with a noval aligment method that Kullback-Leibler divergence loss is performed as probabilistic models $P(o|s) = P(o|t)$.
In implementation stage, $P(o|s) = P(o|t)$ is computed as follows
{
\footnotesize
\begin{equation}
\label{eq1}
    D_{KL}(P(o|t)||P(o|s))=\sum_{j=0}^{J}\sum_{i=0}^{I}P(o_j^i|t,t_j^i)log\frac{P(o_j^i|t,t_j^i)}{P(o_j^i|s,t_j^i)},
\end{equation}
}
where $j$, a hyper-parameter, represents the number of duplicate transcripts.
This is mainly different from \cite{BLSP,BLSPKD}, they only use one simplex prompt `Continue the following text', which may lead to task overfitting phenomenon as described in \cite{salmonn}.

Extensive experiments demonstrate that the resulting Seal model exhibits robust performance as a few-shot learner on two speech understanding tasks, including FSC \cite{fsc} and SLURP \cite{slurp}.
Considering the alignment robustness for various pre-trained language models, phi2 and phi3 pre-trained language models are employed as frozen language model back end.
Experimental results show that the proposed alignment approach achieves the same performance on difference pre-trained language models as speech few-shot learner.

Our contributions are summarized as follows:
\begin{figure*}[t]
\centering
\includegraphics[width=0.75\textwidth]{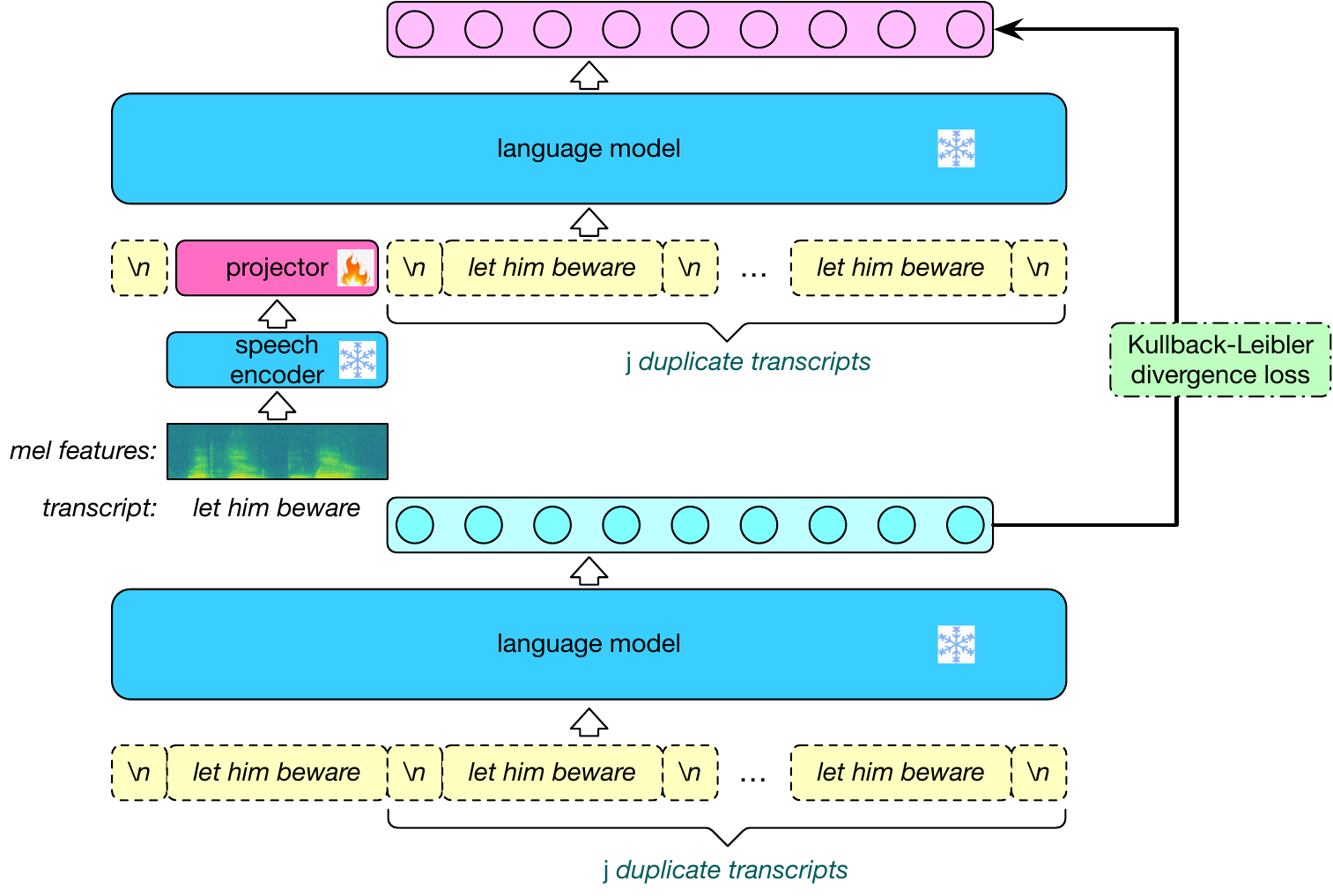} 
\caption{An overview of the Seal model is presented. The Seal model consists of three components: a frozen speech encoder, a trainable projector, and a frozen language model. The Kullback-Leibler divergence loss is used to train the projector, as shown in Equation \ref{eq1}. Specifically, $j+1$ duplicate transcripts, separated by `\textbackslash n', are forwarded to produce the target distribution. This produced distribution then guides the Seal model to simulate the induction head as introduced in \cite{induction}.}
\label{fig1}
\end{figure*}
\begin{itemize}
    \item We present an innovative alignment technique to enhance pre-trained language models to be speech few-shot learners. This involves using Kullback-Leibler divergence loss to align speech features with their corresponding transcripts within the pre-trained language models.
    \item We conduct enrich experiments on various speech understanding tasks for different language models to valid the effectiveness and robustness of our proposed alignment method.
\end{itemize}

\section{Methodology}
This section outlines the proposed \textbf{Seal} model as illustrated in Figure \ref{fig1}, which is mainly composed of three components: a frozen speech encoder, a trainable projector, and a frozen language model. The trainable projector is adjusted according to Equation \ref{eq1}. The specifics of these three components and training method are detailed in the subsequent subsections.

\subsection{Speech encoder}
We utilize the encoder from the Whisper-large-v2 model \cite{whisper} as the speech encoder.
This model is trained using large-scale multilingual and multitask weak supervision data, which is very robust against real-world background sounds \cite{whisperat}.
We use the same pre-processing method that resamples the input speech to 16kHz and then converts the waveform into an 80-channel mel-spectrogram.
The mel-spectrogram is input to transformer blocks \cite{transformer} to produce the speech features.
To reduce the number of tokens occupied by the speech features in the language model, we mask invalid mel-spectrogram during the transformer's forward process. 
Additionally, we use a pooling layer with a stride of 4 to post-process the speech features produced by the transformer's final layer.

\subsection{Projector}
We utilize a trainable projector to align the speech features into word embedding space of language model.
We compute the projector by one linear transformation with a GELU activation \cite{gelu}.
We also incorporate LayerNorm activations \cite{layernorm} and residual connections \cite{residual} to stabilize and accelerate the training process.
Above forward processes are compute as follows:
{
\begin{equation}
\label{eq2}
    w_e=Layernorm(GELU(Layernorm(s_{e}))+s_e),
\end{equation}
}
where $s_e$ denotes the speech feature produced by whisper encoder and $w_e$ denotes the word embedding of corresponding speech features.

We utilize the above linear transformation, which is lightweight and cost-effective, to enable us to explore more alignment methods quickly.
It is also worth noting that there are many candidate architectures, such as Continuous Integrate-and-Fir mechanism \cite{cif}, Q-former \cite{BLIP2} and gated cross-attention \cite{flamingo}, for modeling our trainable projector.
We leave these architectures for future work.

\subsection{Language model}
We utilize the phi-2 model \cite{phi2} as our language model, which is trained on 1.4T tokens from multiple passes on a mixture of synthetic and Web datasets for NLP and coding.
Despite having a 2.7 billion parameter, the phi-2 model demonstrates outstanding reasoning and language understanding capabilities and outperforms models up to 25x larger.
Our project began in December 2023, and at that time, the phi-2 model showed the amazing performance with small size.
Considering the cost-effectiveness, we decided to stick with phi-2 model for all our experiments.
However, in recent years, a series of other models have emerged that could also serve as language models.
To verify the consistency of our alignment method across other models, we also performed experiments on the phi-3 model \cite{phi3}, which is discussed in the experiment section.

\subsection{Training method}

During the training stage, we froze the speech encoder and the language model to maintain the context-learning ability of the language model and to prevent the encoder from losing its ability to capture diverse representations.
Thus, only the projector is tuned to align the speech features in the word embedding space of the language model.
Inspired by \cite{induction}, we perform Kullback-Leibler divergence loss to train the projector as Equation \ref{eq1}.
Specifically, as shown in Figure \ref{fig1}, $j+1$ duplicate transcripts, separated with `\textbackslash n' is forwarded to produce target distribution, and then produced distribution supervise the seal model, which is formulated as Equation \ref{eq1}.
We use $j$ duplicate transcripts to simulate the induction head,
a circuit whose function is to look back over the sequence for previous instances of the current token (call it A), find the token that came after it last time (call it B), and then predict that the same completion will occur again (e.g. forming the sequence [A][B] … [A] → [B]) \cite{induction}.
By simulate induction head, our goal is to ensure that token [B] obtains the same representation, whether it is a text token [B] in the context or a speech feature represented token [B] in the context.
Through the above approach, we equip the pre-train language model with context-learning capabilities for speech.


\section{Experiments}
\label{sec:exp}

    

In this section, we first introduce the pre-training details of the Seal model and then depict the available datasets of speech understanding task in our experiments.
In experiments, we discuss the few-shot learning capabilities of the Seal model on introduced speech understanding tasks. 

\subsection{Pre-train Implementations}
we use two typical speech recognition data to pre-train the proposed Seal model: Common Voice 14.0 English set \cite{commonvoice} and GigaSpeech XL set \cite{Gigaspeech}.
To eliminate the mistaking label in above data and acquire the confidence data, we also use the whisper large-v2 model to clean the data.
We reserve the data whose label is equal to the output text from the whisper model.
Finally, we acquire 6,600+ hours of English data to pre-train the seal model.

During the pre-training, we employ batch size of 1,024, and Adam optimizer with $\beta_1=0.9$ and $\beta_2=0.95$.
We also utilize learning rate warm-up over the first 1,000 steps and constant learning rate of 1e-5 after 1,000 steps.
The total training process takes about 14 days using 4*A100 GPUs on the Phi-2 and 18 days using 4*A100 GPUs on the Phi-2.

\subsection{Details of available datasets}

Considering speech understanding tasks, we utilize two typical datasets: Fluent Speech Commands (FSC) \cite{fsc}, SLURP \cite{slurp}, which includes speech intent classification and speech slot filling.
We detail the aforementioned datasets as follows.

\textbf{FSC:} a intent classification dataset predicts the audio with three slots: action, object and location.
Each slot task one of multiple values: for instance, the `location' slot can take on the values `none', `kitchen', `bedroom', or `washroom'. This dataset refer to the combination of slot values as the intent of the utterance. The dataset has 31 unique intents in total.
We evaluate the performance of this task using accuracy as proposed in \cite{fsc}.
We use inference template of prompt for this task as follows:

\textit{Given the following sentences, firstly, classify the sentences into the following six action categories: (change language, activate, deactivate, increase, decrease, bring), secondly, classify the sentences into the following fourteen object categories: (lights, music, lamp, volume, heat, newspaper, juice, socks, shoes, english, german, chinese, korean, none), finally, classify the sentences into the following four location categories: (kitchen, bedroom, washroom, none). You need to output the annotations in the form of action=X, object=Y, location=Z.\\$\textless speech_0\textgreater\Rightarrow\textless label_0\textgreater$\\...\\
$\textless speech_n\textgreater\Rightarrow\textless label_n\textgreater$\\
$\textless speech_{infer}\textgreater\Rightarrow$},

where $\textless speech_n\textgreater$ and $\textless label_n\textgreater$ represent the speech features and label of the n-th in-context example, respectively. $\textless speech_{infer}\textgreater$ denotes the speech features to predict. Additionally, $\textless speech_n\textgreater$ can be replaced with $\textless text_n\textgreater$ when using a transcript as the in-context example.
\begin{table*}[ht]
    \centering
    \begin{tabular}{l|c|c|c|c|c}
    \hline
    \multirow{2}{*}{Model} & \multirow{2}{*}{0-shot} & \multicolumn{2}{c}{3-shot} & \multicolumn{2}{|c}{5-shot} \\
    \cline{3-6}
    ~ & ~ & Text & Speech & Text & Speech \\
    \hline
    Text+Phi-2 &\textbf{39.23}&78.56&-& 81.03&- \\
    Whisper+Phi-2 &38.54&77.03& -&79.64&- \\
    PSRT & 0.00&10.51&23.71&16.99&19.20\\
    Seal-Phi2 &28.64& 59.75 &68.57&64.54&77.77\\
    Seal-Phi3 &33.30&71.15&77.95&72.34&71.84\\
    Seal-Phi2+KATE &28.64 & 78.24&\textbf{95.76}&83.51&\textbf{96.89}\\
    Seal-Phi3+KATE &33.30 & \textbf{84.77}&95.68&\textbf{88.75}&96.71\\
    \hline
    \end{tabular}
    \caption{Accuracy of FSC intent classification. `Text' denotes using text as in-context examples. `Speech' denotes using speech as in-context examples.}
    \label{tab:fsc}
\end{table*}
\begin{table*}[t]
    \centering
    \begin{tabular}{l|c|c|c|c|c|c|c|c|c|c}
    \hline
    \multirow{3}{*}{Model} & \multicolumn{2}{c}{0-shot} & \multicolumn{4}{|c}{3-shot} & \multicolumn{4}{|c}{5-shot} \\ 
    \cline{2-11}
    ~ & \multirow{2}{*}{Intent} & \multirow{2}{*}{Slot} & \multicolumn{2}{|c}{Intent} &\multicolumn{2}{|c}{Slot} & \multicolumn{2}{|c}{Intent} &\multicolumn{2}{|c}{Slot}\\
    \cline{4-11}
    ~ & ~ & ~ & Text & Speech & Text & Speech & Text & Speech & Text & Speech\\
    \hline
    Text+Phi-2 &2.94&14.66&12.11&-&21.25&-&27.39&-&18.25&-\\
    Whisper+Phi-2 &1.99&12.83&10.48&-&20.23&-&23.86&-&15.78&-\\
    PSRT &0.00&0.00&5.12&4.06&2.52&0.00&7.43&4.48&2.19&0.17\\
    Seal-Phi2 &1.42&8.59&8.77&8.98&20.44&18.91&17.75&15.83&19.98&16.96\\
    Seal-Phi3 &\textbf{17.52}&\textbf{23.98}&18.89&15.34&24.18&23.04&20.62&20.89&23.90&21.48\\
    Seal-Phi2+KATE &1.42&8.59&\textbf{47.01}&52.41&\textbf{42.14}&43.39&\textbf{52.36}&55.76&\textbf{44.22}&45.21\\
    Seal-Phi3+KATE &\textbf{17.52}&\textbf{23.98}&42.97&\textbf{52.77}&37.51&\textbf{45.48}&46.77&\textbf{55.99}&39.58&\textbf{48.09}\\
    \hline
    \end{tabular}
    \caption{Accuracy of slurp intent classification and SLU-F1 of slurp slot filling. `Text' denotes using text as in-context examples. `Speech' denotes using speech as in-context examples.}
    \label{tab:slurp}
\end{table*}

\textbf{SLURP:} a intent classification and a slot filling task predict three levels of semantics: scenario, action, and entities, where scenario and action are called intent classification, entities referred to as `slots' and `values' are identified as slot filling.
This task includes over 18 different scenarios, with 46 defined actions and 55 different entity types.
We evaluate the performance of intent classification using accuracy and slot filling using SLU-F1 as proposed in \cite{slurp}, respectively.
We use inference template of prompt for this task as follows:

\textit{Given the following sentences, firstly, choose the scenario categories of the sentences from the following list: (alarm, audio, calendar, cooking, datetime, email, general, iot, lists, music, news, play, qa, recommendation, social, takeaway, transport, weather), secondly, choose the action categories of the sentences from the following list: (currency, definition, stock, factoid, maths, query, querycontact, sendemail, addcontact, order, remove, createoradd, set, events, locations, movies, quirky, joke, greet, podcasts, music, audiobook, radio, game, likeness, settings, dislikeness, cleaning, hue\_lightchange, hue\_lightdim, hue\_lightoff, wemo\_on, hue\_lightup, wemo\_off, hue\_lighton, post, traffic, taxi, ticket, convert, recipe,volume\_down, volume\_up, volume\_mute), thirdly, annotate given sentences with entities from the following entity list: (movie\_type, device\_type, audiobook\_name, timeofday, definition\_word, media\_type, email\_folder, music\_genre, podcast\_descriptor, news\_topic, business\_name, house\_place, music\_descriptor, coffee\_type, meal\_type, currency\_name, person, transport\_name, order\_type, date, audiobook\_author, food\_type, general\_frequency, app\_name, joke\_type, email\_address, relation, color\_type, artist\_name, business\_type, cooking\_type, ingredient, music\_album, drink\_type, personal\_info, place\_name, time\_zone, podcast\_name, list\_name, radio\_name, game\_name, player\_setting, change\_amount, alarm\_type, playlist\_name, weather\_descriptor, transport\_agency, song\_name, movie\_name, transport\_type, event\_name, time, transport\_descriptor). You need to output the annotations in the form of JSON like this \{``scenario": ``scenario\_value", ``action": ``action\_value", ``entities": [\{"type": ``type\_value0", ``filler": "filler\_value0"\}, \{``type": ``type\_value1", ``filler": ``filler\_value1"\}]\}\\
        $\textless speech_0\textgreater\Rightarrow\textless label_0\textgreater$ \\ ... \\
            $\textless speech_n\textgreater\Rightarrow\textless label_n\textgreater$ \\
            $\textless speech_{infer}\textgreater\Rightarrow$}.







\subsection{Few-shot learning on two speech understanding tasks}

Considering the Few-shot learning on speech understanding tasks, we first detail the models of comparison and then depict the result task-by-task, respectively.

\textbf{Models of comparison:} 1) Text+Phi-2, where we input ground-truth speech transcripts into the Phi-2 model.
This is a ceiling result of used prompt, since no speech recognition model produces ground-truth transcripts.
2) Whisper+Phi-2, where we input transcribed text from Whisper-large-v2 to the Phi-2 model.
This is a particularly strong baseline, as the Whisper model is trained on 680,000 hours of weakly supervised data, which is more than one hundred times the amount used for the Seal model.
3) Pre-trained by Speech Recognition Task (PSRT), where we use pretraining methods similar to those described in \cite{salmonn, qwen-audio} and utilize the same training data as the Seal model.
This is a baseline to examine the in-context learning ability for our proposed alignment method.
4) Seal-Phi2 model, where we use the phi-2 model as pre-trained language models, the same prompt and in-context examples as above models to validate the in-context learning ability of our proposed alignment method.
5) Seal-Phi3 model, where we use the phi-3 model as pre-trained language models, the same prompt and in-context examples as above models to validate alignment method robustness on different pre-trained language models.
6) Seal-Phi2+KATE, where we use the same prompt as above models and we select in-context examples by the k-nearest neighbors algorithm in the embedding space of a speech encoder.
Since the performance of a few-shot learner depend heavily on the choice of in-context example.
We use the above effective strategies in \cite{kate} to judiciously select in-context examples to understand the upper limit of in-context learning capabilities for our proposed method.
7) Seal-Phi3+KATE, where we use the same sample selection strategies as Seal-Phi2+KATE.


\textbf{Results of FSC:}
We show the overall results of the FSC dataset in Table \ref{tab:fsc}. We observe that Seal-Phi2 and Seal-Phi3 obviously outperform PSRT in 0-shot, 3-shot, and 5-shot.
We also observe Seal-Phi2 and Seal-Phi3 still have a certain gap compared to Whisper+Phi2. This is due to a significant difference in the number of training data used.
However, when we used KATE to select the appropriate examples, the performance significantly exceeded the baselines using fixed samples. This indicates that Seal-Phi2 and Seal-Phi3 can sensitively utilize the information provided by the in-context examples for inference. It shows that our approach is highly effective in driving the speech language model to become a few-shot learner.
Despite using only 6,000+ hours dataset to pre-train, the model can quickly adapt to downstream tasks if appropriate in-context examples are selected.

\textbf{Results of SLURP:}
We show the overall results of the SLURP dataset in Table \ref{tab:slurp}. We also observe that Seal-Phi2 and Seal-Phi3 obviously outperform PSRT in 0-shot, 3-shot, and 5-shot.
Due to the relatively large space of intents, the language model has not yet achieved good performance.
When we used KATE to select the appropriate examples, the performance significantly exceeded the baselines.
This indicates that our approach is highly effective in driving the speech language model to become a few-shot learner, whether in simple tasks or complex ones.

Above experiments show that our proposed Seal models can perform a new task once utilizing nicety examples.
And speech as in-context examples tends to achieve better results compared to text as in-context examples.

\section{Discussion}

\subsection{Conclusion}
In this paper, we introduce a novel alignment method that equips the pre-trained language model with context-learning capabilities for speech, where a projector bridges a speech encoder with the pre-trained language model.
Our proposed alignment method is inspired by induction heads \cite{induction}. 
By simulate induction head, our goal is to ensure that token [B] obtains the same representation, whether it is a text token [B] in the context or a speech feature represented token [B] in the context.
Extensive experiments demonstrate that the resulting Seal models exhibit robust performance as a few-shot learner on two speech understanding tasks, including FSC \cite{fsc} and SLURP \cite{slurp}.

\subsection{Future work}
Although, our proposed alignment method shows that it can equip the pre-trained language model with context-learning capabilities for speech.
Its performance still does not surpass the pipeline that the language model joins with the whisper model in zero-shot scenarios.
More data can be used for continued training to improve performance in future work.

In addition, duplicate transcripts are forwarded in the language model to produce a target distribution, which may not be the optimal solution. Continue writing \cite{BLSP,BLSPKD} or replacing duplicate transcripts with random transcripts will be explored in the future.

In particular, more tasks should be considered to benchmark performance in broader dimensions.
This will be the focus of our future work.

\bibliography{aaai24}

\begin{thebibliography}{22}
\providecommand{\natexlab}[1]{#1}

\bibitem[{Abdin et~al.(2024)Abdin, Jacobs, Awan, Aneja, Awadallah, Awadalla, Bach, Bahree, Bakhtiari, Behl, Benhaim, Bilenko, Bjorck, Bubeck, Cai, Mendes, Chen, Chaudhary, Chopra, Giorno, Rosa, Dixon, Eldan, Iter, Garg, Goswami, Gunasekar, Haider, Hao, Hewett, Huynh, Javaheripi, Jin, Kauffmann, Karampatziakis, Kim, Khademi, Kurilenko, Lee, Lee, Li, Liang, Liu, Lin, Lin, Madan, Mitra, Modi, Nguyen, Norick, Patra, Perez-Becker, Portet, Pryzant, Qin, Radmilac, Rosset, Roy, Ruwase, Saarikivi, Saied, Salim, Santacroce, Shah, Shang, Sharma, Song, Tanaka, Wang, Ward, Wang, Witte, Wyatt, Xu, Xu, Yadav, Yang, Yang, Yu, Zhang, Zhang, Zhang, Zhang, Zhang, Zhang, Zhang, and Zhou}]{phi3}
Abdin, M.; Jacobs, S.~A.; Awan, A.~A.; Aneja, J.; Awadallah, A.; Awadalla, H.; Bach, N.; Bahree, A.; Bakhtiari, A.; Behl, H.; Benhaim, A.; Bilenko, M.; Bjorck, J.; Bubeck, S.; Cai, M.; Mendes, C. C.~T.; Chen, W.; Chaudhary, V.; Chopra, P.; Giorno, A.~D.; Rosa, G.~d.; Dixon, M.; Eldan, R.; Iter, D.; Garg, A.; Goswami, A.; Gunasekar, S.; Haider, E.; Hao, J.; Hewett, R.~J.; Huynh, J.; Javaheripi, M.; Jin, X.; Kauffmann, P.; Karampatziakis, N.; Kim, D.; Khademi, M.; Kurilenko, L.; Lee, J.~R.; Lee, Y.~T.; Li, Y.; Liang, C.; Liu, W.; Lin, E.; Lin, Z.; Madan, P.; Mitra, A.; Modi, H.; Nguyen, A.; Norick, B.; Patra, B.; Perez-Becker, D.; Portet, T.; Pryzant, R.; Qin, H.; Radmilac, M.; Rosset, C.; Roy, S.; Ruwase, O.; Saarikivi, O.; Saied, A.; Salim, A.; Santacroce, M.; Shah, S.; Shang, N.; Sharma, H.; Song, X.; Tanaka, M.; Wang, X.; Ward, R.; Wang, G.; Witte, P.; Wyatt, M.; Xu, C.; Xu, J.; Yadav, S.; Yang, F.; Yang, Z.; Yu, D.; Zhang, C.; Zhang, C.; Zhang, J.; Zhang, L.~L.; Zhang, Y.; Zhang, Y.; Zhang, Y.; and Zhou,
  X. 2024.
\newblock {Phi-3 Technical Report: A Highly Capable Language Model Locally on Your Phone}.
\newblock \emph{arXiv}.

\bibitem[{Alayrac et~al.(2022)Alayrac, Donahue, Luc, Miech, Barr, Hasson, Lenc, Mensch, Millican, Reynolds, Ring, Rutherford, Cabi, Han, Gong, Samangooei, Monteiro, Menick, Borgeaud, Brock, Nematzadeh, Sharifzadeh, Binkowski, Barreira, Vinyals, Zisserman, and Simonyan}]{flamingo}
Alayrac, J.-B.; Donahue, J.; Luc, P.; Miech, A.; Barr, I.; Hasson, Y.; Lenc, K.; Mensch, A.; Millican, K.; Reynolds, M.; Ring, R.; Rutherford, E.; Cabi, S.; Han, T.; Gong, Z.; Samangooei, S.; Monteiro, M.; Menick, J.; Borgeaud, S.; Brock, A.; Nematzadeh, A.; Sharifzadeh, S.; Binkowski, M.; Barreira, R.; Vinyals, O.; Zisserman, A.; and Simonyan, K. 2022.
\newblock {Flamingo: a Visual Language Model for Few-Shot Learning}.
\newblock \emph{arXiv}.

\bibitem[{Ardila et~al.(2019)Ardila, Branson, Davis, Henretty, Kohler, Meyer, Morais, Saunders, Tyers, and Weber}]{commonvoice}
Ardila, R.; Branson, M.; Davis, K.; Henretty, M.; Kohler, M.; Meyer, J.; Morais, R.; Saunders, L.; Tyers, F.~M.; and Weber, G. 2019.
\newblock {Common Voice: A Massively-Multilingual Speech Corpus}.
\newblock \emph{arXiv}.

\bibitem[{Ba, Kiros, and Hinton(2016)}]{layernorm}
Ba, J.~L.; Kiros, J.~R.; and Hinton, G.~E. 2016.
\newblock {Layer Normalization}.
\newblock \emph{arXiv}.

\bibitem[{Bastianelli et~al.(2020)Bastianelli, Vanzo, Swietojanski, and Rieser}]{slurp}
Bastianelli, E.; Vanzo, A.; Swietojanski, P.; and Rieser, V. 2020.
\newblock {SLURP: A Spoken Language Understanding Resource Package}.
\newblock \emph{arXiv}.

\bibitem[{Brown et~al.(2020)Brown, Mann, Ryder, Subbiah, Kaplan, Dhariwal, Neelakantan, Shyam, Sastry, Askell, Agarwal, Herbert-Voss, Krueger, Henighan, Child, Ramesh, Ziegler, Wu, Winter, Hesse, Chen, Sigler, Litwin, Gray, Chess, Clark, Berner, McCandlish, Radford, Sutskever, and Amodei}]{gpt3}
Brown, T.~B.; Mann, B.; Ryder, N.; Subbiah, M.; Kaplan, J.; Dhariwal, P.; Neelakantan, A.; Shyam, P.; Sastry, G.; Askell, A.; Agarwal, S.; Herbert-Voss, A.; Krueger, G.; Henighan, T.; Child, R.; Ramesh, A.; Ziegler, D.~M.; Wu, J.; Winter, C.; Hesse, C.; Chen, M.; Sigler, E.; Litwin, M.; Gray, S.; Chess, B.; Clark, J.; Berner, C.; McCandlish, S.; Radford, A.; Sutskever, I.; and Amodei, D. 2020.
\newblock {Language Models are Few-Shot Learners}.
\newblock \emph{arXiv}.

\bibitem[{Catherine et~al.(2022)Catherine, Nelson, Neel, Nicholas, Nova, Tom, Ben, Amanda, Yuntao, Anna, Tom, Dawn, Deep, Zac, Danny, Scott, Andy, Jackson, Liane, Kamal, Dario, Tom, Jack, Jared, Sam, and Chris}]{induction}
Catherine, O.; Nelson, E.; Neel, N.; Nicholas, J.; Nova, D.; Tom, H.; Ben, M.; Amanda, A.; Yuntao, B.; Anna, C.; Tom, C.; Dawn, D.; Deep, G.; Zac, H.-D.; Danny, H.; Scott, J.; Andy, J.; Jackson, K.; Liane, L.; Kamal, N.; Dario, A.; Tom, B.; Jack, C.; Jared, K.; Sam, M.; and Chris, O. 2022.
\newblock In-context Learning and Induction Heads.
\newblock \url{https://transformer-circuits.pub/2022/in-context-learning-and-induction-heads/index.html}.

\bibitem[{Chen et~al.(2021)Chen, Chai, Wang, Du, Zhang, Weng, Su, Povey, Trmal, Zhang, Jin, Khudanpur, Watanabe, Zhao, Zou, Li, Yao, Wang, Wang, You, and Yan}]{Gigaspeech}
Chen, G.; Chai, S.; Wang, G.; Du, J.; Zhang, W.-Q.; Weng, C.; Su, D.; Povey, D.; Trmal, J.; Zhang, J.; Jin, M.; Khudanpur, S.; Watanabe, S.; Zhao, S.; Zou, W.; Li, X.; Yao, X.; Wang, Y.; Wang, Y.; You, Z.; and Yan, Z. 2021.
\newblock {GigaSpeech: An Evolving, Multi-domain ASR Corpus with 10,000 Hours of Transcribed Audio}.
\newblock \emph{arXiv}.

\bibitem[{Chu et~al.(2023)Chu, Xu, Zhou, Yang, Zhang, Yan, Zhou, and Zhou}]{qwen-audio}
Chu, Y.; Xu, J.; Zhou, X.; Yang, Q.; Zhang, S.; Yan, Z.; Zhou, C.; and Zhou, J. 2023.
\newblock {Qwen-Audio: Advancing Universal Audio Understanding via Unified Large-Scale Audio-Language Models}.
\newblock \emph{arXiv}.

\bibitem[{Dong and Xu(2020)}]{cif}
Dong, L.; and Xu, B. 2020.
\newblock CIF: Continuous Integrate-And-Fire for End-To-End Speech Recognition.
\newblock In \emph{ICASSP 2020 - 2020 IEEE International Conference on Acoustics, Speech and Signal Processing (ICASSP)}, 6079--6083.

\bibitem[{Gong et~al.(2023)Gong, Khurana, Karlinsky, and Glass}]{whisperat}
Gong, Y.; Khurana, S.; Karlinsky, L.; and Glass, J. 2023.
\newblock {Whisper-AT: Noise-Robust Automatic Speech Recognizers are Also Strong General Audio Event Taggers}.
\newblock \emph{arXiv}.

\bibitem[{He et~al.(2015)He, Zhang, Ren, and Sun}]{residual}
He, K.; Zhang, X.; Ren, S.; and Sun, J. 2015.
\newblock {Deep Residual Learning for Image Recognition}.
\newblock \emph{arXiv}.

\bibitem[{Hendrycks and Gimpel(2016)}]{gelu}
Hendrycks, D.; and Gimpel, K. 2016.
\newblock {Gaussian Error Linear Units (GELUs)}.
\newblock \emph{arXiv}.

\bibitem[{Li et~al.(2023)Li, Li, Savarese, and Hoi}]{BLIP2}
Li, J.; Li, D.; Savarese, S.; and Hoi, S. 2023.
\newblock {BLIP-2: Bootstrapping Language-Image Pre-training with Frozen Image Encoders and Large Language Models}.
\newblock \emph{arXiv}.

\bibitem[{Liu et~al.(2021)Liu, Shen, Zhang, Dolan, Carin, and Chen}]{kate}
Liu, J.; Shen, D.; Zhang, Y.; Dolan, B.; Carin, L.; and Chen, W. 2021.
\newblock {What Makes Good In-Context Examples for GPT-\$3\$?}
\newblock \emph{arXiv}.

\bibitem[{Lugosch et~al.(2019)Lugosch, Ravanelli, Ignoto, Tomar, and Bengio}]{fsc}
Lugosch, L.; Ravanelli, M.; Ignoto, P.; Tomar, V.~S.; and Bengio, Y. 2019.
\newblock {Speech Model Pre-training for End-to-End Spoken Language Understanding}.
\newblock \emph{arXiv}.

\bibitem[{Marah et~al.(2023)Marah, Jyoti, Sebastien, Caio, Weizhu, Allie, Ronen, Sivakanth, Suriya, Mojan, Piero, Yin, Yuanzhi, Anh, Gustavo, Olli, Adil, Shital, Michael, Harkirat, Adam, Xin, Rachel, Philipp, Cyril, and Yi}]{phi2}
Marah, A.; Jyoti, A.; Sebastien, B.; Caio, C. T.~M.; Weizhu, C.; Allie, D.~G.; Ronen, E.; Sivakanth, G.; Suriya, G.; Mojan, J.; Piero, K.; Yin, T.~L.; Yuanzhi, L.; Anh, N.; Gustavo, d.~R.; Olli, S.; Adil, S.; Shital, S.; Michael, S.; Harkirat, S.~B.; Adam, T.~K.; Xin, W.; Rachel, W.; Philipp, W.; Cyril, Z.; and Yi, Z. 2023.
\newblock Phi-2: The surprising power of small language models.
\newblock \url{https://www.microsoft.com/en-us/research/blog/phi-2-the-surprising-power-of-small-language-models/}.

\bibitem[{Radford et~al.(2023)Radford, Kim, Xu, Brockman, McLeavey, and Sutskever}]{whisper}
Radford, A.; Kim, J.~W.; Xu, T.; Brockman, G.; McLeavey, C.; and Sutskever, I. 2023.
\newblock {Robust Speech Recognition via Large-Scale Weak Supervision}.
\newblock \emph{arXiv}.

\bibitem[{Tang et~al.(2023)Tang, Yu, Sun, Chen, Tan, Li, Lu, Ma, and Zhang}]{salmonn}
Tang, C.; Yu, W.; Sun, G.; Chen, X.; Tan, T.; Li, W.; Lu, L.; Ma, Z.; and Zhang, C. 2023.
\newblock {SALMONN: Towards Generic Hearing Abilities for Large Language Models}.
\newblock \emph{arXiv}.

\bibitem[{Vaswani et~al.(2017)Vaswani, Shazeer, Parmar, Uszkoreit, Jones, Gomez, Kaiser, and Polosukhin}]{transformer}
Vaswani, A.; Shazeer, N.; Parmar, N.; Uszkoreit, J.; Jones, L.; Gomez, A.~N.; Kaiser, L.; and Polosukhin, I. 2017.
\newblock {Attention Is All You Need}.
\newblock \emph{arXiv}.

\bibitem[{Wang et~al.(2023)Wang, Liao, Huang, Lu, Wu, Liu, Zong, and Zhang}]{BLSP}
Wang, C.; Liao, M.; Huang, Z.; Lu, J.; Wu, J.; Liu, Y.; Zong, C.; and Zhang, J. 2023.
\newblock {BLSP: Bootstrapping Language-Speech Pre-training via Behavior Alignment of Continuation Writing}.
\newblock \emph{arXiv}.

\bibitem[{Wang et~al.(2024)Wang, Liao, Huang, and Zhang}]{BLSPKD}
Wang, C.; Liao, M.; Huang, Z.; and Zhang, J. 2024.
\newblock {BLSP-KD: Bootstrapping Language-Speech Pre-training via Knowledge Distillation}.
\newblock \emph{arXiv}.

\end{thebibliography}


\end{document}